\documentclass[11pt,a4paper]{article}
\usepackage[hyperref]{acl2021}
\usepackage{times}
\usepackage{latexsym}

\usepackage[section]{placeins}
\usepackage{amsmath}
\usepackage{booktabs}
\usepackage{multirow}
\usepackage{color}
\usepackage{graphicx}
\usepackage{enumitem}
\setlist{nolistsep}
\usepackage{array}
\usepackage{makecell}
\usepackage{pifont}

\usepackage{microtype}

\aclfinalcopy 
\def\aclpaperid{1028} 

\title{Exploiting Position Bias for Robust Aspect Sentiment Classification}

\author{Fang Ma\Thanks{ Fang Ma and Chen Zhang contribute equally to this work. The order is determined alphabetically.}, Chen Zhang\footnotemark[1], Dawei Song\Thanks{ Dawei Song is the corresponding author.} \\
   School of Computer Science, Beijing Institute of Technology \\
   Beijing, China \\
   \texttt{\{mfang,czhang,dwsong\}@bit.edu.cn}
}

\date{}

\begin{document}
\maketitle
\begin{abstract}
Aspect sentiment classification (ASC) aims at determining sentiments expressed towards different aspects in a sentence. While state-of-the-art ASC models have achieved remarkable performance, they are recently shown to suffer from the issue of robustness. Particularly in two common scenarios: when domains of test and training data are different (out-of-domain scenario) or test data is adversarially perturbed (adversarial scenario), ASC models may attend to irrelevant words and neglect opinion expressions that truly describe diverse aspects. To tackle the challenge, in this paper, we hypothesize that position bias (i.e., the words closer to a concerning aspect would carry a higher degree of importance) is crucial for building more robust ASC models by reducing the probability of mis-attending. Accordingly, we propose two mechanisms for capturing position bias, namely position-biased weight and position-biased dropout, which can be flexibly injected into existing models to enhance representations for classification. Experiments conducted on out-of-domain and adversarial datasets demonstrate that our proposed approaches largely improve the robustness and effectiveness of current models.\footnote{The code and preprocessed data are available at \url{https://github.com/BD-MF/POS4ASC}.}
\end{abstract}

\section{Introduction}

Aspect sentiment classification (ASC) is an important sub-task of sentiment classification. It aims to identify the sentiment polarity (i.e., negative, neutral, or positive) of a specified aspect in a sentence. Take ``\textit{Great food but the service was bad.}'' as an example. For aspects \textit{food} and \textit{service}, their corresponding sentiment polarities are \textit{positive} and \textit{negative}, respectively.

A challenge in ASC is how to model semantic relations between aspect terms and their contexts, which requires an ASC model to be only sensitive to the sentiment words actually depicting the target aspect terms. Although previous ASC models~\citep{tang2016aspect,li2018transformation,zhang2019aspect,xu2019bert,wang2020relational,tang2020dependency} have achieved promising results by modeling complex interactions between aspects and contexts, these models have recently been shown to suffer from the lack of robustness~\citep{xing2020tasty}. The issue is particularly severe in two scenarios: 1) out-of-domain (O.O.D.) scenario: ASC models that perform well on training data often fail to generalize to test data in another domain; 2) adversarial (Adv.) scenario: ASC models can be easily fooled by small adversarially perturbed inputs, e.g. synonymous word substituted ones. To our best knowledge, none of current ASC models have been targeted at alleviating the robustness issue in above-mentioned two scenarios.

\begin{figure}[t]
\centering
\resizebox{0.47\textwidth}{!}{
\begin{tabular}{ccc}
\toprule  
\textbf{Scenario} & \textbf{Example} & \textbf{Pred./Lb.}     \\ 
\midrule
I.D. & \colorbox{orange!0.589}{\strut Great} \colorbox{orange!2.35}{\strut food} \colorbox{orange!1.90}{\strut but} \colorbox{orange!0.19}{\strut the} \colorbox{orange!0.10}{\strut \underline{service}} \colorbox{orange!0.00}{\strut was} \colorbox{orange!100.00}{\strut bad} \colorbox{orange!2.45}{\strut !} & neg./neg. \\
\midrule
O.O.D. & \colorbox{orange!0.00}{\strut The} \colorbox{orange!12.39}{\strut \underline{battery}} \colorbox{orange!5.31}{\strut has} \colorbox{orange!15.13}{\strut never} \colorbox{orange!13.29}{\strut worked} \colorbox{orange!100.00}{\strut well}
\colorbox{orange!0.00}{\strut .} & pos./neg. \\
\midrule
Adv. & \colorbox{orange!100.00}{\strut Awful} \colorbox{orange!4.86}{\strut food} \colorbox{orange!2.45}{\strut but} \colorbox{orange!0.52}{\strut the} \colorbox{orange!0.70}{\strut \underline{service}} \colorbox{orange!0.15}{\strut was} \colorbox{orange!2.29}{\strut great} \colorbox{orange!0.00}{\strut !} & neg./pos. \\
\bottomrule
\end{tabular}}
\caption{An illustration of how an ASC model IAN~\citep{ma2017interactive} might fail. Gradient saliency maps~\citep{simonyan2014deep} with respect to the embedding of each word in I.D., O.O.D., and Adv. scenarios, along with the model predictions (Pred.) and corresponding ground truth labels (Lb.), are provided. \underline{Underlined} words are aspects.}
 \label{fig1}    
\end{figure}

To fill this gap, inspired by a recent finding that highlighting words close to a target aspect (termed as \textit{position bias}) would boost in-domain (I.D.) effectiveness of a model~\citep{zhang2019syntax}, we hypothesize that such position bias is also crucial for a robust ASC model in O.O.D. and Adv. scenarios. Figure~\ref{fig1} shows an illustrative example of an ASC model that fails in the two scenarios due to mis-attending. In contrast, with position bias, a model tends to focus more on words nearer to the aspect, thus reducing the probability of mis-attending. Concretely, we propose two mechanisms: position-biased weight and position-biased dropout. The former assigns an inductive weight to each word according to its position proximity to the aspect. The latter gives each word a probability of being reserved (or dropped out) according to its proximity relation to the aspect. In doing so, position-biased weight degrades the significance of words that are not close enough to the aspect, while position-biased dropout will drop those likely irrelevant words at high probabilities. 

\begin{figure}[t]
    \centering
    \includegraphics[width=0.47\textwidth]{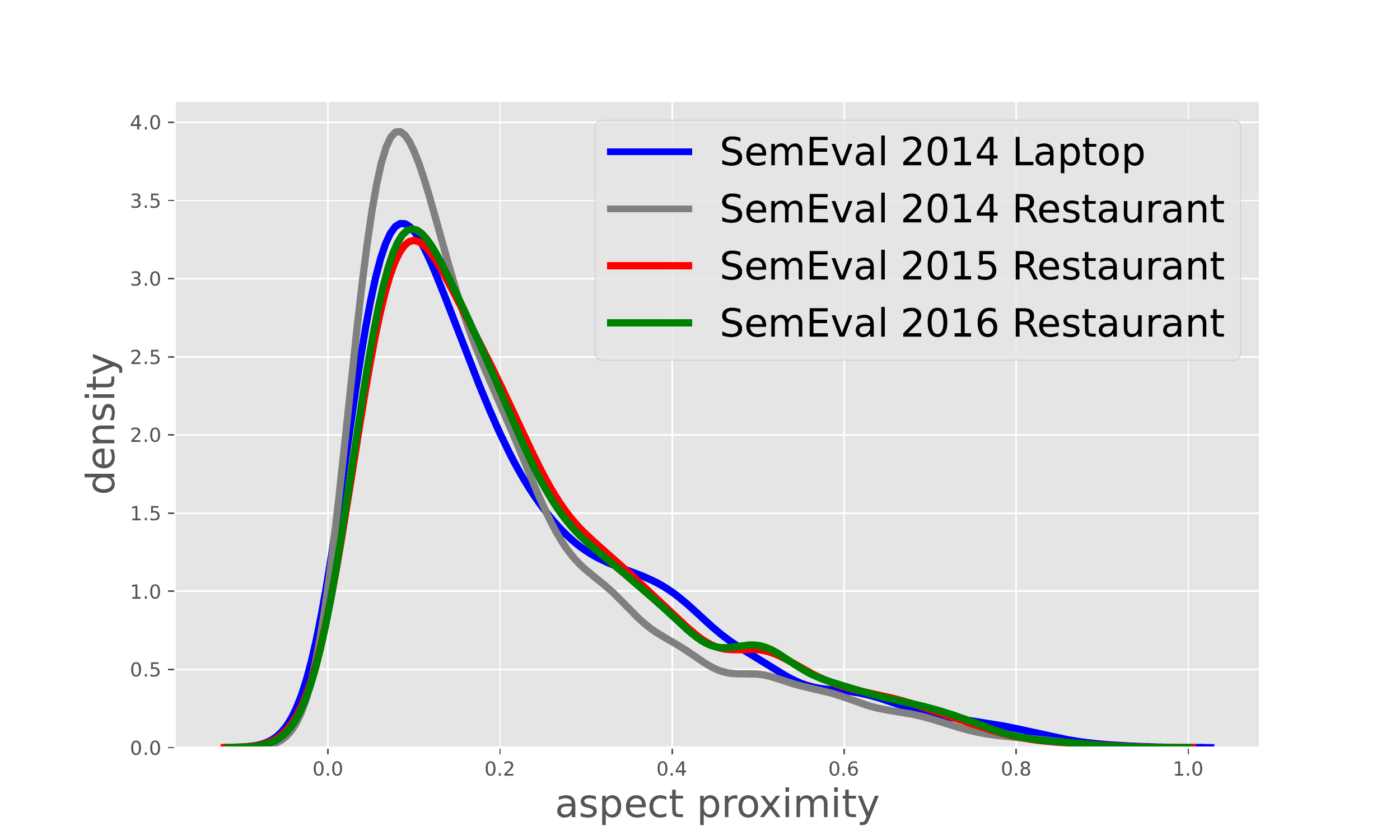}
    \caption{Aspect proximity distributions of different benchmarks, visualized with the kernel density estimation.}
    \label{fig2}
\end{figure}

Essentially, position bias is quantitatively evidenced in commonly used benchmarks. With annotated aspect-opinion pairs offered by~\citet{fan2019target}, we can calculate position proximity between any pair of aspect and opinion (in short, aspect proximity) in a sentence. The aspect proximity is computed by dividing the relative distance between a pair of aspect and opinion by the length of the corresponding sentence. Therefore, we can plot aspect proximity distributions of these benchmarks with kernel density estimation, as shown in Figure~\ref{fig2}. These distributions indicate the aspect proximity is small at a high probability, thereby position bias is a reasonable inductive bias.

Extensive experiments are conducted on SemEval and ARTS datasets~\citep{xing2020tasty}. The results show that incorporating the proposed position bias mechanisms would lead to more robust ASC models in both out-of-domain and adversarial scenarios. Furthermore, in terms of flexibility, the proposed methods can be easily adapted by subsequent models.

\section{Capturing Position Bias}

This section describes the proposed position-biased weight and position-biased dropout for capturing position bias. Formally, an $n$-word sentence containing a target $m$-word aspect term is formulated as $S=\{w_0, w_1,\dots,w_\gamma,w_{\gamma+1},\dots,w_{n-1}\}$, where $\gamma$ denotes the start index of the aspect term. By resorting to either a pre-trained word embedding~\citep{bengio2003neural} or a pre-trained language model~\citep{devlin2019bert}, we can represent the sentence as $V=\{e_0,e_1,\dots,e_\gamma,e_{\gamma+1},\dots,e_{n-1}\}$. We can then use the position-biased weight and dropout to refine $V$ and generate an enhanced representation, denoted as $E=\{h_0,h_1,...h_{\gamma},h_{\gamma+1},...,h_{n-1}\}$. $E$ can then be incorporated into the model, i.e., any further structures for the model will be built upon $E$, instead of $V$, to predict sentiment polarities associated with diverse aspects.

\paragraph{Position-biased Weight}

Generally, the sentiment polarity of an aspect term is determined by its context, which are the words around the aspect term~\citep{zhang2019syntax}. Thus we can leverage relative position information to calculate weights of context words, with the aim to degrade the significance of those words that are far away from the aspect. Position-biased weight, denoted as $p_i\in(0,1)$, is computed as:
\begin{equation}
    p_i=\left\{
    \begin{array}{lcr}
    1-\frac{\gamma-i}{n-m} & & {0 \leq i < \gamma}\\
    \frac{1}{n-m} & & {\gamma \leq i < \gamma+m}\\
    1-\frac{i-\gamma-m+1}{n-m} & & {\gamma+m \leq i < n}\\
    \end{array}
    \right.
    \label{eq1}
\end{equation}
Then we compute $h_i$ as: $h_i=p_i\cdot e_i$.

\paragraph{Position-biased Dropout}

Dropout~\citep{srivastava2014dropout,sennrich2016edinburgh} randomly sets elements in a feature vector to zeros. The word-level dropout can model semantic and syntactic compositionality and reduce input redundancy~\citep{iyyer2015deep}. Motivated by this idea, we give each word a probability of being reserved according to its position proximity to the aspect. The aim is to preserve those words that are close enough to the aspect and drop the rest. The probability that the $i$-th word will be preserved can be computed as:
\begin{equation}
z_i\sim \mathrm{Bernoulli}(p_i)
\end{equation}
where $\mathrm{Bernoulli}(p_i)$ denotes that $z_i$ equals to 1 with $p_i$ and equals to 0 with $1-p_i$. The $i$-th word is dropped out if $z_i$ is 0. Likewise, $h_i$ can be attained by multiplying $z_i$ and $e_i$.

\section{Experiments}

\paragraph{Datasets}

To empirically evaluate a model's O.O.D. robustness, we conduct experiments on two domains from SemEval 2014~\citep{pontikisemeval} dataset (\textsc{SemEval}): one is laptop (\textsc{SemEval-Lap}) and the other is restaurant (\textsc{SemEval-Rest}). For Adv. robustness, we experiment with the Aspect Robustness Test dataset (\textsc{Arts})~\citep{xing2020tasty}, which is derived from the SemEval 2014 dataset. Instances in \textsc{Arts} are generated with three adversarial strategies. These strategies enrich the test set from 638 to 1,877 for the laptop domain (\textsc{Arts-Lap}), and from 1,120 to 3,530 for the restaurant domain (\textsc{Arts-Rest}). Note that each domain from \textsc{SemEval} consists of separate training and test sets, while each domain from \textsc{Arts} only contains a test set. Since \textsc{SemEval} dataset does not come with development sets, 150 instances from the training set in each domain are randomly selected to form the development set. Table \ref{tab2} shows the statistics of the datasets.

\begin{table}[ht]
\centering
\resizebox{0.42\textwidth}{!}{
\begin{tabular}{ccccc}
\toprule
\multicolumn{2}{c}{\textbf{Dataset}}       & \textbf{\# pos.} & \textbf{\# neu.} & \textbf{\# neg.} \\ 
\midrule
\multirow{3}{*}{\textsc{SemEval-Lap}}     & train & 930      & 433     & 800      \\
\cmidrule{2-5}
                            & test  & 341      & 169     & 128      \\
                            \cmidrule{2-5}
                            & dev   & 57       & 27      & 66       \\ 
                            \midrule
\multirow{3}{*}{\textsc{SemEval-Rest}} & train & 2,094     & 579     & 779      \\
\cmidrule{2-5}
                            & test  & 728      & 196     & 196      \\
\cmidrule{2-5}
                            & dev   & 70       & 54      & 26       \\
\midrule
\textsc{Arts-Lap}                    & test  & 883      & 407     & 587      \\ 
\midrule
\textsc{Arts-Rest}                  & test  & 1,953     & 473     & 1,104     \\ 
\bottomrule
\end{tabular}}
\caption{Statistics of datasets.}
\label{tab2}  
\end{table}

\paragraph{Target Models}

We conduct experiments on a wide range of existing models for a comprehensive study on whether position bias is beneficial. Specifically, we examine these models' performance before and after injecting the position bias, in terms of position-biased weight (\texttt{pos-wt}) and dropout (\texttt{pos-dp}) individually.

The target models include: (a) \textbf{LSTM} \citep{tang2016effective} uses the last hidden state vector of the LSTM to predict sentiment. (b) \textbf{LSTM-Attn} \citep{wang2016attention} applies an attention-based LSTM on the concatenation of the aspect and word embeddings. (c) \textbf{IAN} \citep{ma2017interactive} interactively learns attentions between context words and aspect terms. (d) \textbf{MemNet} \citep{tang2016aspect} applies attention multiple times on word memories, and the output of the last attention is used for prediction. While the original work utilizes word embeddings as memories, we instead choose to add a layer of bidirectional LSTM upon embeddings for more abstractive memories. (e) \textbf{AOA} \citep{huang2018aspect} introduces an attention-over-attention based network to model interaction between aspects and contexts. (f) \textbf{RoBERTa}~\citep{dai2021does} is a strong baseline with an MLP built upon the pooled feature induced with RoBERTa~\citep{liu2019roberta}. 

\paragraph{Implementation Details}

In all our experiments, the 300-dimensional GloVe~\citep{pennington2014glove} is leveraged to initialize the input embedding. All parameters of models are initialized with uniform distributions. During all experiments, a bidirectional LSTM is adopted if necessary instead of a unidirectional one. If a model takes advantage of the attention mechanism, then dot product based attention is employed. In case a model has hidden states, the dimensionality of hidden states is set to 300. The batch size is 64. We use Adam \citep{kingma2014adam} as the optimizer with a learning rate of 10\textsuperscript{-3}. The coefficient of L2 regularization is 10\textsuperscript{-5}. For experiments with RoBERTa as the input embedding~\citep{liu2019roberta}, things may change. The dimensionality of hidden states is 768. The learning rate is 10\textsuperscript{-5}, while the regularization is rather removed.

\paragraph{Evaluation Metrics}

For O.O.D., models are trained separately on one domain and evaluated on another. For Adv., models are trained on the \textsc{SemEval} dataset and tested on the \textsc{Arts} counterpart. For every test, a model is trained on the I.D. training set, selected on the I.D. development set, and tested on the O.O.D. or Adv. test set. The experimental results are obtained by averaging 5 runs with random initialization, and we adopt Accuracy and macro-averaged F1 scores as evaluation metrics.

\begin{table*}[t]
\centering 
\resizebox{0.99\textwidth}{!}{
\begin{tabular}{lccccccccc}
\toprule
\multirow{3}{*}{\textbf{Model}} &
  \multicolumn{4}{c}{\textbf{\textsc{Lap}}} &
  \multicolumn{4}{c}{\textbf{\textsc{Rest}}} \\ \cmidrule(l){2-9} 
 &
  \multicolumn{2}{c}{\textbf{O.O.D.}} &
  \multicolumn{2}{c}{\textbf{Adv.}} &
  \multicolumn{2}{c}{\textbf{O.O.D.}} &
  \multicolumn{2}{c}{\textbf{Adv.}} \\ \cmidrule(l){2-9} 
 &
  \textbf{Acc.} & \textbf{F1} &
  \textbf{Acc.} & \textbf{F1} &
  \textbf{Acc.} & \textbf{F1} &
  \textbf{Acc.} & \textbf{F1} \\ \midrule
LSTM       
&71.02 & 52.15 & 49.49 & 43.91 
&60.60 & 53.25 & 53.34 & 41.99 \\

\quad w/ \texttt{pos-dp}  
& 71.48\small{$\uparrow$\textbf{0.46}} & 50.98\small{$\downarrow$\textbf{1.17}} 
& 50.74\small{$\uparrow$\textbf{1.25}} & 44.38\small{$\uparrow$\textbf{0.47}} 
& 63.39\small{$\uparrow$\textbf{2.79}} & 58.57\small{$\uparrow$\textbf{5.32}} 
& 53.57\small{$\uparrow$\textbf{0.23}} & 42.11\small{$\uparrow$\textbf{0.12}} \\

\quad w/ \texttt{pos-wt}  
& 72.96\small{$\uparrow$\textbf{{\textcolor{red}{1.94}}}}  & 55.88\small{$\uparrow$\textbf{{\textcolor{red}{3.73}}}} 
& 55.50\small{$\uparrow$\textbf{{\textcolor{red}{6.01}}}} & 50.03\small{$\uparrow$\textbf{{\textcolor{red}{6.12}}}} 
& 66.33\small{$\uparrow$\textbf{{\textcolor{red}{5.73}}}} & 60.21\small{$\uparrow$\textbf{{\textcolor{red}{6.96}}}}
& 59.03\small{$\uparrow$\textbf{{\textcolor{red}{5.69}}}} & 48.20\small{$\uparrow$\textbf{{\textcolor{red}{6.21}}}} \\

\midrule

LSTM-Attn    
& 71.61 & 53.61 & 51.33 & 46.11 
& 62.85 & 54.97 & 58.45 & 49.65 \\

\quad w/ \texttt{pos-dp}    
& 71.34\small{$\downarrow$\textbf{0.27}} & 52.49\small{$\downarrow$\textbf{1.12}} 
& 53.76\small{$\uparrow$\textbf{2.43}} & 48.47\small{$\uparrow$\textbf{2.36}} 
& 65.24\small{$\uparrow$\textbf{2.39}} & 59.07\small{$\uparrow$\textbf{4.10}} 
& 58.64\small{$\uparrow$\textbf{0.19}} & 47.22\small{$\downarrow$\textbf{2.43}} \\

\quad w/ \texttt{pos-wt}    
& 72.84\small{$\uparrow$\textbf{{\textcolor{red}{1.23}}}} & 56.18\small{$\uparrow$\textbf{{\textcolor{red}{2.57}}}}
& 58.53\small{$\uparrow$\textbf{{\textcolor{red}{7.20}}}} & 53.54\small{$\uparrow$\textbf{{\textcolor{red}{7.43}}}} 
& 68.90\small{$\uparrow$\textbf{{\textcolor{red}{6.05}}}} & 64.48\small{$\uparrow$\textbf{{\textcolor{red}{9.51}}}}
& 64.80\small{$\uparrow$\textbf{{\textcolor{red}{6.35}}}} & 55.34\small{$\uparrow$\textbf{{\textcolor{red}{5.69}}}} \\

\midrule

IAN    
& 72.09 & 54.44 & 52.91 & 47.54 
& 63.82 & 55.20 & 57.75 & 48.12 \\
\quad w/ \texttt{pos-dp}    
& 70.95\small{$\downarrow$\textbf{1.14}} & 51.63\small{$\downarrow$\textbf{3.08}} 
& 52.04\small{$\downarrow$\textbf{0.87}} & 45.87\small{$\downarrow$\textbf{1.67}} 
& 63.57\small{$\downarrow$\textbf{0.25}} & 56.81\small{$\uparrow$\textbf{{\textcolor{red}{1.61}}}} 
& 56.89\small{$\downarrow$\textbf{0.86}} & 46.90\small{$\downarrow$\textbf{1.22}} \\
\quad w/ \texttt{pos-wt}    
& 72.86\small{$\uparrow$\textbf{{\textcolor{red}{0.77}}}} & 54.88\small{$\uparrow$\textbf{{\textcolor{red}{0.44}}}} 
& 56.03\small{$\uparrow$\textbf{{\textcolor{red}{3.12}}}} & 50.30\small{$\uparrow$\textbf{{\textcolor{red}{2.76}}}} 
& 62.45\small{$\downarrow$\textbf{1.37}} & 55.95\small{$\uparrow$\textbf{0.75}} 
& 63.49\small{$\uparrow$\textbf{{\textcolor{red}{5.74}}}} & 54.04\small{$\uparrow$\textbf{{\textcolor{red}{5.92}}}} \\
\midrule

MemNet    
& 70.66 & 52.07 & 52.00 & 46.50 
& 57.84 & 51.15 & 55.30 & 46.67 \\
\quad w/ \texttt{pos-dp}    
& 69.93\small{$\downarrow$\textbf{0.73}} & 53.37\small{$\uparrow$\textbf{1.30}} 
& 53.54\small{$\uparrow$\textbf{1.54}} & 47.93\small{$\uparrow$\textbf{1.43}} 
& 61.94\small{$\uparrow$\textbf{{\textcolor{red}{4.10}}}} & 54.49\small{$\uparrow$\textbf{3.34}} 
& 57.31\small{$\uparrow$\textbf{2.01}} & 45.23\small{$\downarrow$\textbf{1.44}} \\
\quad w/ \texttt{pos-wt}    
& 70.67\small{$\uparrow$\textbf{\textcolor{red}{0.01}}} & 54.14\small{$\uparrow$\textbf{{\textcolor{red}{2.07}}}} 
& 56.04\small{$\uparrow$\textbf{{\textcolor{red}{4.04}}}} & 49.64\small{$\uparrow$\textbf{{\textcolor{red}{3.14}}}} 
& 61.35\small{$\uparrow$\textbf{3.51}} & 54.85\small{$\uparrow$\textbf{{\textcolor{red}{3.70}}}} 
& 61.10\small{$\uparrow$\textbf{{\textcolor{red}{5.80}}}} & 51.49\small{$\uparrow$\textbf{{\textcolor{red}{4.82}}}} \\

\midrule

AOA    
& 71.63 & 52.65 & 52.16 & 46.78 
& 63.73 & 57.00 & 58.19 & 49.02 \\
\quad w/ \texttt{pos-dp}    
& 72.30\small{$\downarrow$\textbf{0.67}} & 53.73\small{$\uparrow$\textbf{1.08}} 
& 53.56\small{$\uparrow$\textbf{1.40}} & 48.18\small{$\uparrow$\textbf{1.40}} 
& 65.33\small{$\uparrow$\textbf{1.60}} & 58.31\small{$\uparrow$\textbf{1.31}} 
& 56.24\small{$\downarrow$\textbf{1.95}} & 45.63\small{$\downarrow$\textbf{3.39}} \\
\quad w/ \texttt{pos-wt}    
& 72.61\small{$\uparrow$\textbf{{\textcolor{red}{0.98}}}} & 56.54\small{$\uparrow$\textbf{{\textcolor{red}{3.89}}}} 
& 59.07\small{$\uparrow$\textbf{{\textcolor{red}{6.91}}}} & 54.92\small{$\uparrow$\textbf{{\textcolor{red}{8.14}}}} 
& 66.87\small{$\uparrow$\textbf{{\textcolor{red}{3.14}}}} & 62.02\small{$\uparrow$\textbf{{\textcolor{red}{5.02}}}}
& 64.35\small{$\uparrow$\textbf{{\textcolor{red}{6.16}}}} & 54.62\small{$\uparrow$\textbf{{\textcolor{red}{5.60}}}} \\
\midrule

RoBERTa
& 83.16 & 72.99 & 73.57 & 69.26 
& 77.62 & 71.34 & 79.08 & 71.79 \\
\quad w/ \texttt{pos-dp}    
& 81.98\small{$\downarrow$\textbf{1.18}} & 70.81\small{$\downarrow$\textbf{{2.18}}} 
& 69.98\small{$\downarrow$\textbf{3.59}} & 65.35\small{$\downarrow$\textbf{3.91}} 
& 75.61\small{$\downarrow$\textbf{2.01}} & 68.00\small{$\downarrow$\textbf{3.34}} 
& 77.81\small{$\downarrow$\textbf{{1.27}}} & 69.37\small{$\downarrow$\textbf{2.42}} \\
\quad w/ \texttt{pos-wt}    
& 83.43\small{$\uparrow$\textbf{{\textcolor{red}{0.27}}}} & 74.08\small{$\uparrow$\textbf{{\textcolor{red}{1.09}}}} 
& 75.72\small{$\uparrow$\textbf{{\textcolor{red}{2.15}}}} & 72.09\small{$\uparrow$\textbf{{\textcolor{red}{2.83}}}} 
& 79.40\small{$\uparrow$\textbf{{\textcolor{red}{1.78}}}} & 74.44\small{$\uparrow$\textbf{{\textcolor{red}{3.10}}}}
& 79.47\small{$\uparrow$\textbf{{\textcolor{red}{0.39}}}} & 73.10\small{$\uparrow$\textbf{{\textcolor{red}{1.31}}}} \\
 \bottomrule
\end{tabular}}
 
\caption{Robustness results (\%). O.O.D. on \textsc{Lap} or \textsc{Rest} denotes a model is trained in current domain (\textsc{Lap} or \textsc{Rest}) and tested on another (\textsc{Rest} or \textsc{Lap}). Adv. denotes a model is trained in a domain and tested on its ARTS counterpart. Furthermore, w/ \texttt{pos-dp} means a model with position-biased dropout. w/ \texttt{pos-wt} means a model with position-biased weight. The small number next to each performance score indicates either performance improvement ($\uparrow$) or drop ($\downarrow$) compared with the original model without using position bias, and those highlighted in \textcolor{red}{red} are the best-performing ones among two variants.}
\label{tab3}
\end{table*}

\paragraph{In-domain Generalization Results}

Table~\ref{ID_Result} shows the I.D. performance of the LSTM on both laptop and restaurant domains, which exhibits incorporating position bias does not harm, if this is the case, a model's generalization on I.D. test sets that much. On the contrary, position bias, especially with position-biased weight, can boost I.D. performance. 

\begin{table}[ht]
\centering
\resizebox{0.40\textwidth}{!}{
\begin{tabular}{lcccc}
\toprule
\multirow{2}{*}{\textbf{Model}} &
  \multicolumn{2}{c}{\textbf{\textsc{Lap I.D.}}} & \multicolumn{2}{c}{\textbf{\textsc{Rest I.D.}}} \\ \cmidrule(l){2-5}
  & \textbf{Acc.} & \textbf{F1} & \textbf{Acc.} & \textbf{F1} \\ \midrule
  LSTM  & 67.15 & 60.57 & 74.57 & 62.14 \\
 \quad w/ \texttt{pos-dp} & 67.34 & 60.27 & 74.23 & 61.55 \\
 \quad w/ \texttt{pos-wt} & 68.78 & 62.42 & 76.34 & 64.85 \\ \bottomrule
\end{tabular}}
\caption{I.D. results (\%) of LSTM on \textsc{Lap} and \textsc{Rest}.}
\label{ID_Result}  
\end{table}

\paragraph{Robustness Results}

The robustness results are shown in Table~\ref{tab3}. We can see that performance of LSTM drops drastically, compared to I.D. performance, on O.O.D. and Adv. test sets, indicating the importance of studying the robustness issue. Our proposed two position bias mechanisms improve the target models' O.O.D. and Adv. performance in most cases. With position-biased dropout, F1 scores of models are improved by up to 5.32 pp on O.O.D. test sets, and 2.36 pp on Adv. test sets, though the efficacy of the position-biased dropout seems not stable across different target models and settings. In contrast, the impact of position-biased weight is much more prominent. With position-biased weight, Accuracy scores of models can be enhanced by up to 6.05 pp and 7.20 pp on O.O.D. and Adv. test sets, respectively. Further, F1 scores of models are improved by up to 9.51 pp and 8.14 pp on O.O.D. and Adv. test sets. 

A highlight is that experimental results with RoBERTa as well exhibit the benefit of position bias, yet with caveats. Although pre-trained language models like RoBERTa are subject to positional encodings, such absolute position information is not enough to model relative position relations between aspect terms and contexts. Therefore, position bias matters during fine-tuning pre-trained language models for robust ASC performance. However, we observe that position-biased dropout is not an appropriate choice for pre-trained language models.

\paragraph{Case Study}

To understand the effect of position bias, we conduct a case study on the two robustness scenarios, as shown in Table~\ref{position_impact}. Specifically, we visualize the attention scores separately offered by an ASC model LSTM-Attn with and without position-biased weight method and trained on \textsc{SemEval-Rest}. 

We can observe that before applying position bias, the model attends irrelevant words and fails in both scenarios. Specifically, in both cases, the model mis-attends to irrelevant opinion expressions. After injecting position bias, the attention scores become more accurate and the model attends to the correct opinion spans. 

\begin{table}[ht]
\centering
\resizebox{0.48\textwidth}{!}{
\begin{tabular}{ccc}
\toprule
w/ \texttt{pos-wt} & Example  \\ 
\midrule
\ding{56} & \makecell[c]{
\colorbox{orange!1.377}{\strut The} 
\colorbox{orange!2.892}{\strut \underline{price}}
\colorbox{orange!35.40}{\strut is}
\colorbox{orange!17.303}{\strut reasonable}
\colorbox{orange!4.807}{\strut although}\\
\colorbox{orange!1.907}{\strut the}
\colorbox{orange!0.931}{\strut quality}
\colorbox{orange!77.616}{\strut is}
\colorbox{orange!100.0}{\strut poor}
\colorbox{orange!0.0}{\strut .}} \\
\midrule
\ding{52} & 
\makecell[c]{\colorbox{orange!0.000}{\strut The} 
\colorbox{orange!0.240}{\strut \underline{price}}
\colorbox{orange!0.809}{\strut is}
\colorbox{orange!100.0}{\strut reasonable}
\colorbox{orange!0.407}{\strut although}\\
\colorbox{orange!0.0}{\strut the}
\colorbox{orange!0.413}{\strut quality}
\colorbox{orange!0.315}{\strut is}
\colorbox{orange!4.725}{\strut poor}
\colorbox{orange!0.363}{\strut .}} \\
\midrule
\ding{56} & \colorbox{orange!100.00}{\strut Awful} \colorbox{orange!0.039}{\strut food} \colorbox{orange!0.006}{\strut but} \colorbox{orange!0.0}{\strut the} \colorbox{orange!0.013}{\strut \underline{service}} \colorbox{orange!0.044}{\strut was} \colorbox{orange!31.5}{\strut great} \colorbox{orange!19.87}{\strut !} \\
\midrule
\ding{52} & \colorbox{orange!50.958}{\strut Awful} \colorbox{orange!1.799}{\strut food} \colorbox{orange!0.4216}{\strut but} \colorbox{orange!0.084}{\strut the} \colorbox{orange!0.0}{\strut \underline{service}} \colorbox{orange!0.419}{\strut was} \colorbox{orange!100.0}{\strut great} \colorbox{orange!0.593}{\strut !} \\
\bottomrule
\end{tabular}}
\caption{\textbf{Case study.} The \underline{underlined} words are aspects. The top two rows are O.O.D. examples, while the bottom two are Adv. examples. \ding{56} and \ding{52} refers to without and with \texttt{pos-wt} respectively.}
\label{position_impact}    
\end{table}

\section{Related Work}

\paragraph{Fine-grained Sentiment Analysis} 

ASC falls in the broad scope of fine-grained sentiment analysis. While ASC is basically formulated as determining sentiment polarity of a given aspect in a sentence~\citep{tang2016aspect,tang2016effective,wang2016attention,chen2017recurrent,huang2018aspect,li2018transformation,xu2019bert,zhang2019syntax,zhang2019aspect,wang2020relational,tang2020dependency}, there is an emergent trend that treating fine-grained sentiment as an opinion triplet extraction task~\citep{peng2020knowing,zhang2020multi,wu2020grid}. Recently, the robustness of ASC models becomes a critical issue that urges researchers to pay more attention on improving the robustness of ASC models~\citep{xing2020tasty}. Our work is the first work to enhance the universal robustness of ASC models by capturing position bias. On another note, we believe opinion triplet extraction is exposed to the similar robustness issue, which should be explored in the near future.

\paragraph{Robustness in NLP}

Broadly, there are two kinds of robustness in NLP, i.e., O.O.D. and Adv. robustness. O.O.D. robustness in NLP has attracted extensive attention in recent work~\citep{ng2020ssmba,hendrycks2020pretrained,xie2020n}. In terms of O.O.D. robustness, they often use the cross-domain setting to evaluate models~\citep{benson2020assessing}. Previous work mainly focuses on how to minimize the domain discrepancy and how to improve the feature adaptability of models~\citep{rietzler2020adapt,ye2020feature}. On the other hand, adversarial learning becomes the main method used to improve Adv. robustness of models~\citep{xing2020tasty}. Prior methods consider using semantic operations, such as synonym replacement, random insertion, random swap, and random deletion to augment data~\citep{wei2019eda}. Other methods involve adding extra text~\citep{wallace2019universal} and replacing sentences with semantically similar sentences~\citep{ribeiro2018semantically}. Our work goes beyond the two forms of robustness and aims to achieve universal robustness for ASC with position bias.

\section{Conclusion and Future Work}

In this work, we find that state-of-the-art ASC models suffer from the issue of robustness, particularly in two scenarios: i) out-of-domain scenario, and ii) adversarial scenario. To address the issue, we propose a simple yet effective inductive bias that should be incorporated, that is, position bias. We proposed two mechanisms to capture position bias, namely position-biased weight and position-biased dropout. They are injected into existing models to enhance the representation. Extensive experiments demonstrate that the proposed methods can largely improve the models' robustness. The results verify our hypothesis that position bias is beneficial for building more robust ASC models.

The work shall be improved in the following two facets: i) Since the approach of incorporating position bias is straightforward yet naive, especially for pre-trained language models, it is meaningful to consider a nicely designed architecture to inject position bias in a more elegant manner. ii) It has been shown that position bias for ASC is highly correlated with the syntactic structure of the sentence. Hence, syntax can likewise be explored to enhance the robustness of ASC models. 

\section*{Acknowledgements}

This work is supported by the National Key Research and Development Program of China (grant No. 2018YFC0831704) and the Natural Science Foundation of China (grant No. U1636203).

\bibliographystyle{acl_natbib}
\bibliography{anthology,acl2021}


\end{document}